\documentclass[conference]{IEEEtran}

\usepackage{cite}
\usepackage{amsmath,amssymb,amsfonts}
\usepackage{mathtools}
\usepackage{algorithmic}
\usepackage{graphicx}
\usepackage{textcomp}
\usepackage{xcolor}
\usepackage{url}
\usepackage{tikz}
\def\BibTeX{{\rm B\kern-.05em{\sc i\kern-.025em b}\kern-.08em
    T\kern-.1667em\lower.7ex\hbox{E}\kern-.125emX}}

\newcommand\copyrighttext{%
  \footnotesize \textcopyright~2026 IEEE. Personal use of this material is permitted.
  Permission from IEEE must be obtained for all other uses, in any current or future
  media, including reprinting/republishing this material for advertising or promotional
  purposes, creating new collective works, for resale or redistribution to servers or
  lists, or reuse of any copyrighted component of this work in other works.}
\newcommand\copyrightnotice{%
\begin{tikzpicture}[remember picture,overlay]
\node[anchor=south,yshift=8pt] at (current page.south) {\parbox{\dimexpr\textwidth\relax}{\copyrighttext}};
\end{tikzpicture}%
}

\begin{document}

\author{
\IEEEauthorblockN{Lakshya Garg, Deep Narayan Mishra, Swapnil Yadav, Haoan Wang,\\Sujal Alugubelli, Karthik Kumaran, Anupriya Sharma}
\IEEEauthorblockA{Walmart Global Tech, Sunnyvale, CA, USA\\
\{lakshya.garg, deep.mishra, swapnil.yadav, haoan.wang,\\sujal.alugubelli, karthik.kumaran, anupriya.sharma0\}@walmart.com}
}

\title{Demand Transfer Estimation at Scale via Restricted Logit Modeling}

\maketitle
\copyrightnotice
\begin{abstract}
Item demand forecasting is an integral component of store assortment optimization (SAO). Existing literature focuses on learning a suitable customer choice model and using this model to determine the value of an objective function (i.e. expected demand) with respect to an assortment proposal. However, for large item universe with many categories, this approach can prove inefficient, needing a separate demand forecast for every possible item assortment. An alternate approach to SAO exists whereby we combine the efficiency of forecasting item demand independently, while at the same time applying adjustments to the independent forecasts that account for the relations between item demand and the availability of other similar items on the shelf. \newline

Central to this approach is the estimation of Demand Transfer (DT) coefficients. These DT coefficients represent the percent of a particular target item’s (Item that the customer walked in the store to buy) demand that is redirected to each other item in the universe should the target item be removed from the shelf. We introduce an approach that allows us to compute these DT coefficients on large item universes (assortments having 1 million+ items). Experiments on data as well as historical transaction data for multiple locations within categories demonstrate that when certain reasonable assumptions about substitution behavior are satisfied, our procedure is able to accurately estimate underlying DT coefficients and lead to improvements in demand forecasting.

\end{abstract}


\section{Introduction}
With the widespread adoption of eCommerce retail platforms, it has become more important than ever for physical stores to make data driven decisions to remain competitive. One of the single most important decisions a brick-and-mortar retailer can make is Store Assortment Optimization (SAO), the decision of which items to stock on a particular shelf, and how many of each item to carry[4, 7]. Such systems typically involve the maximization of some objective function (typically a function of revenue, demand, or some other business metrics) with respect to some constraints (e.g. physical shelf space). Among the most important inputs to an SAO framework are the individual item demand forecasts. These predictions are used to estimate the expected total revenue that corresponds to a particular proposed assortment. \newline

However, due to physical constraints, item demand forecasting within a brick-and-mortar context can prove difficult, because the demand profiles of individual items within a shelf are often dependent on the availability of other similar items that may serve a similar need. Such correlations between the items mean we cannot model the time series corresponding to an items' demand independently of the other items we intend to put on the shelf alongside it. This phenomenon of demand cannibalization and customer preference of buying other items is also a part of Demand transference (DT) and is amplified when on-shelf assortments are consistently modified [12]. \newline

Below we talk about the mathematical formulation of DT. Consider two items $i, j$ in an assortment $A = \{i, j\}$ that both fulfill a certain customer need  (i.e., they are substitutable products as discussed in Section III-A (Item Substitution)). Let us have an estimate of the expected total demand $D_{A}$ of this assortment, given by
\begin{equation}
\mathbb{E}[D_A] = \mathbb{E}[D_{i,A}] + \mathbb{E}[D_{j,A}],
\end{equation}
where $D_{i,A}$ and $D_{j,A}$ are random variables (with known distributions)  representing the demand of items $i$ and $j$, respectively, under assortment $A$. We assume that a customer who is interested in purchasing item $i$ would also be willing to purchase item $j$ - as they are substitutable items - with probability $\rho_{ij}: \coloneqq \mathbb{P}(i \to j)$ in the case that $i$ is not available  for purchase but all other items in the universe are available.

Suppose that we wish to compute the expected total demand $\mathbb{E}[D_{A'}]$ of a modified assortment $A'$, consisting of only item $j$. This means that if item $i$ is removed from the shelf, we cannot compute the expected demand of the modified assortment by simply subtracting and using equation 1
\begin{equation}
\mathbb{E}[D_{j,A'}] \neq \mathbb{E}[D_{j,A}] = \mathbb{E}[D_A] - \mathbb{E}[D_{i,A}],
\end{equation}
because we cannot consider all of the demand that item $i$ would have received under assortment $A$ to be totally lost. Some portion of it, $\mathbb{P}(i \to j)$, is expected to transfer to item $j$, provided that item $j$ is available for purchase, as is the case in assortment $A'$. So the correct formulation of $\mathbb{E}[D_{A'}]$ should be:
\begin{equation}
\mathbb{E}[D_{A'}] = \rho_{ij}\,\mathbb{E}[D_{i,A}] + \mathbb{E}[D_{j,A}]
\end{equation}

Here $\rho_{ij}$ ( $\mathbb{P}(i \to j)$) captures the additional demand that would be transferred to item $j$. This $\rho_{ij}$ is called \emph{Demand Transference (DT)} from item $i$ to item $j$.

Most approaches to solve demand of similar items have used a Choice Model as a component in a broader function that describes Revenue or some suitable objective function, which is then optimized as seen in Abdallah and Vulcano [3], Fisher and Vaidyanathan [7], this is computationally unsuitable for large item universe and for our downstream SAO use case of identifying how many units of each SKU should be put on the shelf. For our use case explicit transfer coefficients between items are needed, which some works such as Blanchet et al. [5], Şimşek and Topaloglu [14] have dealt with using a Markov Chain model.  However, the methods used by Blanchet et al. [5], Şimşek and Topaloglu [14]  to estimate transition probabilities suffer from scaling and stability issues for very large item universes.  \emph {The work that we propose in this paper is a modification of the Markov Chain model to estimate explicit transfer coefficients between items leveraging a substitution framework that makes the original algorithm scalable for a huge item universe (1 million+ items)}.
This scalability is of utmost importance because:
\begin{itemize}
    \item As with most modern retailers, whether small or large, assortments typically consist of at least ten thousands of items.
    \item Assortment sizes continue to grow as consumers demand an ever-increasing variety of choices especially in General merchandise (GM) categories .
\end{itemize}

As our proposed method is scalable across huge item sets we can extend DT to all categories, thereby enabling any retailer big or small to make more data driven assortment decisions.

\section{RELATED WORK}
One way to account for DT effects within the demand modeling process is to control for them by separately modeling an item's demand $D_{i,a}$ under each potential assortment $a \in \{X \mid X \subseteq U\}$, where $U$ denotes the (universe) set of items under analysis. This formulation has been adopted in much of the relevant literature, in which a general revenue function $r : 2^{U} \to \mathbb{R}$ is maximized by identifying an optimal subset of items from the universe. DT is incorporated through the inclusion of a customer choice model that provides the probability that a customer will purchase item $j$ given that a subset $S \subseteq U$ is available on the shelf.

The other way of solving the DT problem has focused on identifying a suitable choice model and then estimating its parameters (Abdallah and Vulcano [3], Fisher and Vaidyanathan [7]), or directly estimating its function if it is a non-parametric model, as in Arias et al. [6]. These approach as are inadequate for our purposes for two main reasons.
\begin{enumerate}
    \item A large retail operation may contain thousands of categories, each with very large item counts, which makes such an approach computationally impractical. For a single category $U$, performing an optimization over all possible item-availability combinations would require as many as $|U| \cdot 2^{|U|}$ separate time-series forecasts.   
        \item  Most possible assortments will not have been historically implemented in any stores due to the large item count, leading to a lack of data to make reliable demand forecasts corresponding to these assortments. This unreliability in the demand forecasts for historically unseen combinations could prevent the discovery of novel optimal assortments during the optimization step.
        \item Finally, the above approaches only consider the question of which items to make available on the shelf and not how many of each of those selected items should be put on shelf. As our optimization engine must answer both of these questions we require explicit DT coefficients.      
\end{enumerate}

We also look at the work done by Fisher and Vaidyanathan [7], which considers estimation of all inter-item DT probabilities (there referred to as substitution probabilities) with respect to a certain attribute. However, the focus of that work is primarily in improving downstream revenues; the probability estimates are computed alongside the primary demand estimates in the optimization of a joint likelihood function on observed sales data. Furthermore there is no validation procedure dedicated to the quality of the probability estimates themselves; they are treated as latent parameters in a global revenue optimization, which as highlighted above is not suitable for our use case. \newline

One important line of research originated in the work of Blanchet et al. [5] and was continued by Şimşek and Topaloglu [14]. These authors consider analysis of customer DT behavior as a sequence of transitions between states (corresponding to items) in a Markov Chain, whose transition matrix may be interpreted as our desired DT matrix $\rho$. However, the method for estimation of the transition probabilities provided by Blanchet et al. [5] requires non-sparsity in historically available assortments, which is not the case in our setting, and while this is avoided in the expectation-maximization algorithm provided by Şimşek and Topaloglu [14], their approach suffers from numerical instability at scale since the involved linear systems can become inconsistent for large item universes, and computation is slow when many possible subsets are considered [14]. One workaround would be to estimate the parameters of multiple smaller such models, each on a disjoint partition of the overall item universe. However, in our context, a natural way of partitioning an item category is unavailable due to overlapping need states, in our use case need state is defined by the purpose the item is fulfilling for the customer. Thus, a the original implementation of Markov Chain model was unsuitable in our setting. \newline

A key result shown in Blanchet et al. [5] is that a Markov Chain choice model becomes mathematically equivalent to a Multinomial Logit (MNL) model when the Markov Chain’s transition matrix has rank one---that is, when every item has the same pattern of transition probabilities to all other items in the assortment. This equivalence is important because the MNL model is far more scalable and computationally efficient than a general Markov Chain model. Therefore, we leverage this result---together with reasonable assumptions about customer demand---to estimate the parameters of an MNL model. Once the MNL parameters are estimated, we apply a simple transformation to interpret them as the transition probabilities of an equivalent Markov Chain choice model. \newline

We also note that when an item is removed from the shelf, it is often the case that a portion of its demand does \emph{not} transfer to any other item. This corresponds to the \emph{no-purchase option} [14], which can be interpreted as demand transferring to a special item $\phi$ that is always available for purchase but is not considered part of any assortment. We estimate the quantities $\rho_{i\phi}$ for all $i \in U$ (U denotes the item universe) using business-rule-based deterministic logic from historical item availability and observed demand retention behavior, and then allocate the remaining fraction of transferable demand across other items in proportion to their standard DT coefficients.

In addition to the assortment optimization and choice-modeling literature, product cannibalization has been extensively studied in marketing science and retail analytics. Mason and Milne (1994) [19] investigated methods for identifying cannibalization effects within product line extensions and multi-brand strategies, highlighting the importance of quantifying demand redistribution between related products. More recently, Xu (2025) [20] proposed a clustering-based approach for detecting product cannibalization using price effects, demonstrating the increasing use of data-driven methods to characterize substitution behavior. These studies reinforce the importance of understanding demand interactions between products, although most focus on cannibalization detection rather than estimating explicit item-to-item demand transfer coefficients at the scale considered in this work.

\section{ITEM SUBSTITUTION FRAMEWORK}

In the following section we go over our item substitution framework which when used in conjunction to our proposed modification of MNL algorithm makes the original algorithm scalable.

\subsection{Item Substitution}
Item substitution is a symmetric, anti-reflexive[1], and non-transitive relation corresponding to the similarity of two items from a customer’s perspective. High substitutability between two items means that in general, customers are willing to purchase one item in the others place, or consider them functionally equivalent. This suggests that all substitutable items with respect to a given target item can serve as an implicit characterization of the target item within the same need state. \newline

This characterization offers an additional advantage: together with the known equivalence between a rank-one Markov chain transition matrix and an MNL model [5], we can seamlessly integrate the parameters of a choice model to produce the final DT coefficients. As our choice model assumes IIA (Independence of Irrelevant Alternatives discussed in subsection 4.1), we directly incorporate the set of substitutable items as a condition on the marginal purchase probabilities. \newline

According to assumption 3 (discussed in subsection 4.1), conditioning the probability distribution on a specific set S being the set of substitutes for a removed item i is equivalent to conditioning on the need state of i. This relationship provides the required conditional probabilities to characterize demand-transfer behavior. When taken together, these assumptions imply that modeling the item-to-item DT coefficient $\rho_{ij}$ reduces to two steps: \newline

\begin{enumerate}
	\item Identifying the subset $S \subseteq U$ of items that are substitutable for item $i$
        \item Modeling customer choice as the probability distribution of an arbitrary customer purchasing among only the items in S that are currently available.
\end{enumerate}

\subsection{Calculation of Substitute Scores}
Prior work on estimating the substitutability of items has focused on utilizing customer purchase data to determine the substitutability of an item pair. Typically, when the same customer is observed to have repeatedly purchased two different items under similar conditions, this pattern is taken as a positive signal that the items are substitutes.

We develop a comprehensive framework that addresses the complexity of real-world substitution patterns. The theoretical foundation for modeling substitution builds on discrete choice theory and customer utility maximization. We assume that customers have underlying preferences that can be modeled probabilistically.

Our approach to detect substitutability between items is to introduce a so-called \textit{Substitution Score} for every pair of items. The substitution relation introduced previously corresponds to an indicator $s_{ij} \in \{0,1\}$ for a given item pair $(i,j)$, which equals 1 if and only if the two products serve a similar need-state, and thus should be considered as candidates for demand transfer for each-other.

We define $s_{ij}$, $\forall (i,j) \in \mathcal{U}$ as follows:
\begin{equation}
s_{ij} = I(\theta_{ij} \geq \tau); \quad \theta_{ij}, \tau \in \mathbb{R}
\end{equation}

where $\theta_{ij}$ represents the result of a decision variable for items $i,j$ and $\tau$ represents a threshold value with $U$ being the item universe. It is this $\theta_{ij}$ which we refer to as our substitute score, and is a continuous variable expressing the degree to which $i,j$ are related. If $\theta_{ij}$ is above our chosen threshold, we can conclude that $i,j$ are substitutes of one another.

One challenge we faced was that no single method of computing such a decision variable $\theta_{ij}$ and threshold $\tau$ proved applicable for all pairs of items within our universe. Different item categories exhibit vastly different operational scales and purchase frequencies. For example, using in-store co-purchase patterns proved very useful for complement detection for pairs with high purchase frequencies, but less so for items with highly seasonal demand. This meant that we had to employ a variety of different methodologies for computing the substitute scores for different sets of item pairs, depending on which signals were appropriate to a particular context.

Below we talk about all different methodologies used for calculating substitute scores between item pairs. 

\subsection{Store Yule's Q Methodology}

The most reliable kind of data with respect to estimating substitute scores for an DT context is the historical store customer purchase data itself. Representing the ultimate source of ground-truth (the actual purchase decisions of customers). Methodologies that use this kind of data can only be relied upon when we have sufficiently many purchases of both items (target and substitutable item) under question to decide whether their purchase patterns are (statistically) significantly associated. This condition is satisfied by many item pairs in our universe, and the methodology we use to measure their substitutability is called the \textit{Store Yule's Q} (YQ) score.

The first step in calculating a pair's Store Yule's Q score is to compute its \textit{Odds Ratio} (OR). The OR is a measure of association between two events. It is a metric ranging from 0 to $\infty$ and is defined as the relative probability between event 1 given event 2 and the probability of event 1 given event 2 does not happen. Mathematically, the OR between events A and B is:

\begin{equation}
\text{OR} = \frac{n_{A \cap B}}{n_{B \cap A^C}} \div \frac{n_{A \cap B^C}}{n_{(A \cup B)^C}} = \frac{n_{A \cap B} \times n_{(A \cup B)^C}}{n_{B \cap A^C} \times n_{A \cap B^C}}
\label{eq:odds_ratio}
\end{equation}

where $n_x$ is the number of times event $x$ occurred. 

In the context of our retail data, we apply this OR calculation to customers and baskets separately to determine the association and distinction between same customer purchasing different items vs same basket containing different items.

Odds ratios can then be standardized into a Yule's Q number using the following formula:

\begin{equation}
\text{YQ} = \frac{\text{OR} - 1}{\text{OR} + 1}
\label{eq:yules_q}
\end{equation}

which ranges from $-1$ to $1$.

\subsection{Customer Odds Ratio}

One of the key inputs into the substitute Yule's Q metric is the Customer Odds Ratio. For items A and B, it is calculated as follows:

\begin{equation}
\text{OR}_{\text{cust}} = \frac{n\text{Cust}_{A \cap B} \times n\text{Cust}_{(A \cup B)^C}}{n\text{Cust}_{B \cap A^C} \times n\text{Cust}_{A \cap B^C}}
\label{eq:customer_or}
\end{equation}

where $n\text{Cust}_x$ indicates the number of customers that have purchased $x$.

\subsection{Basket Odds Ratio}

Although the Customer odds ratio is a useful metric for detecting items with a high association from a customer purchase perspective, its utility is hampered by the fact that items that are highly dissimilar but are often purchased together (i.e., complements such as bread and butter) will also tend to have high association along with true substitutes. In order to filter these complement pairs out, we utilize another metric called a Basket Odds Ratio, which measures the association of two items from a cart/basket perspective. The formula for this basket odds ratio is:

\begin{equation}
\text{OR}_{\text{basket}} = \frac{n\text{Basket}_{A \cap B} \times n\text{Basket}_{(A \cup B)^C}}{n\text{Basket}_{B \cap A^C} \times n\text{Basket}_{A \cap B^C}}
\label{eq:basket_or}
\end{equation}

where $n\text{Basket}_x$ indicates the number of baskets that contain $x$ in the data.

Before a substitute odds ratio is created, an adjustment is made to the customer odds ratio:

\begin{equation}
\text{OR}_{\text{Cust-Adj}} = \min(\text{OR}_{\text{Cust}}, \text{OR}_{\text{Basket}} + 10)
\label{eq:customer_adjustment}
\end{equation}
\footnotetext{The additive constant of 10 was introduced in equation number 9 as a regularization parameter to limit the influence of extreme customer odds ratios.}

This adjustment is used to reduce the effect of extremely large customer odds ratios on final substitute scores, particularly when the basket odds ratio is also large.

A substitute odds ratio is then calculated from the adjusted customer and basket odds ratios using the following formula:

\begin{equation}
\text{OR}_{\text{subs}} = \frac{\text{OR}_{\text{Cust-Adj}}}{\text{OR}_{\text{Basket}} + 1}
\label{eq:substitute_or}
\end{equation}

By substituting Equation 10 in Equation 6 a Substitute Yule's Q score is created. Therefore we get a final continuous substitute score (which corresponds to our decision variable $\theta_{ij}$ for all $i,j \in U$ mentioned in the methodology) in the range $[-1,1]$

\begin{table}[!t]
\centering
\caption{Substitution score Interpretation}
\label{tab:customer_odds}
\begin{tabular}{|c|p{7.5cm}|}
\hline
\textbf{Score Range} & \textbf{Description} \\
\hline
$> 0$ &
Customers that buy one product are more likely to buy the other product than someone who has not. \\
\hline
$= 0$ &
Buying one of the products does not change the likelihood a customer will buy the other. \\
\hline
$< 0$ &
Customers that purchase one of the products are less likely to buy the other. \\
\hline
\end{tabular}
\end{table}

\subsection{Threshold Selection}
Throughout this work, items with substitution scores greater than 0.6 are considered substitutes. The threshold was selected based on empirical evaluation conducted during model development. Lower thresholds increased substitute-set coverage but introduced a larger number of weak substitution relationships that frequently corresponded to complementary rather than substitutable products. Conversely, higher thresholds improved precision but reduced substitute-set coverage, limiting measurable demand-transfer effects.

\subsection{Low selling / new items}
New/low selling items often lack the historical sales data required to generate reliable substitution scores through Store Yule's Q. To address this cold-start/low velocity item problem, we employ SBERT-based similarity models that leverage product attributes, including item descriptions, brand, and price, to rank the most similar items within the assortment and infer potential substitute relationships.

\section{MNL FRAMEWORK}
In the following section we go over our proposed MNL Framework.

\subsection{Modeling Demand Transfer Coefficients}
As our DT proportions are used as multiplicative factors of expected demand forecasts, they can be interpreted as probabilities. The estimation of $\boldsymbol{\rho} := \{\rho_{ij}\}_{i,j \in U}$ amounts to estimating a set of conditional probabilities \(\rho_{ij} := \mathbb{P}(B_j \mid I_i)\), where $B_j$ denotes the event that an arbitrary customer ultimately purchases item $j$, and $I_i$ denotes the event that the customer initially intended to purchase item $i$, and that every item in $U$ was available except for $i$.

In our characterization of DT effects via the stochastic coefficient matrix $\boldsymbol{\rho} \in [0,1]^{|U| \times |U|}$, the value of each element $\rho_{ij}$ is determined by two phenomena:
\begin{enumerate}
    \item \textbf{Item Loyalty:} For certain items, when they become unavailable
    for purchase, some customers are unwilling to switch to any alternative.
    Also referred as the no-purchase option.
    
    \item \textbf{Item switching:} If a customer is willing to switch to an 
    alternative item, their preferences over the remaining available items are
    distributed according to the need state fulfilled by the item they originally 
    intended to purchase.
\end{enumerate}

With respect to these phenomena, we make the following modeling assumptions: \newline
Assumption 1: Item loyalty and item switching preferences are independent factors. That is, \[D_{ij} = (1 - \rho_{i\phi}) \cdot \rho_{ij},\] where $D_{ij}$ represents the final percentage of demand transferred from item $i$ to item $j$ after accounting for the no-purchase option. \newline

We only have access to historical (observational) sales data corresponding to purchase outcomes under a limited scope of historical assortments. This presents the issue that many demand transfer scenarios for which we must produce predictions have never occurred in the past, and are difficult to estimate directly. Furthermore, we do not have access to the underlying preferences of the historical customers, but only their final purchase decisions in a variety of availability scenarios. This presents additional difficulty in characterizing a purchase as a result of demand transferred to another item, or as a result of initial interest in the item. \newline

For these two above reason, we adopt a statistical modeling approach for customer choice that obeys an IIA assumption, and estimate its parameters using the available observational purchase/availability data. The result of this approach is a set of parameters, one for each item in the universe, that can be interpreted as that item’s "mean utility" across customers [14, 17]. By virtue of the IIA assumption, the utility parameters can be passed through a simple transformation to yield a probability distribution (corresponding to probability of purchase) over all items in the universe [17]. \newline

we take into account the need state fulfilled by a particular item when determining its DT coefficients from the choice model parameters as distribution over the entire item universe cannot by itself be considered as a representation of demand transfer. So given a customer had some initial interest in the initial purchase item it should indicate some degree of interest in similar substitutable items and no interest at all in items of an entirely different nature. Given this we make our second assumption. \newline

Assumption 2: Item switching preferences are entirely determined by the underlying need state associated with an initial interest in the deleted item. That is, \[\rho_{ij} = \mathbb{P}(B_j \mid I_i) = \mathbb{P}(B_j \mid N_i)\] where $N_i$ denotes the event that the customer has the need state associated with item $i$, and that all items in the universe except $i$ are available. A consequence of assumption 2 is that an item’s Demand Transfer profile depends not on the item itself, but on that item’s need state. \newline

Assumption 3: The need state corresponding to a customer's initial interest in a particular item is fully characterized by the set of items that are sufficiently substitutable for it. That is,
\[\mathbb{P}(B_j \mid I_i)
    = \mathbb{P}(B_j \mid N_i)
    = \mathbb{P}\big( B_j \,\big|\, \text{only } \{ m \in U \mid \sigma_{mi} = 1 \} )
\]
where $\sigma_{mi}$ is an indicator function equal to $1$ whenever $m$ and $i$ are substitutable items. Moreover, these substitutable items are the only items to which a customer initially interested in purchasing item $i$ would consider switching.  \newline

The last Assumption 3, is important for justifying our approach, it states that item--switching behavior is statistically dependent on the initial preference only through the set of items that a customer would consider as substitutes. If three items $i$, $j$, and $k$ belong in the same substitute set \(S \subseteq \{i,j,k\}\), then, $\rho_{ij}$ the probability that a customer initially interested in item $i$ will instead purchase item $j$ is equal to the probability that an arbitrary customer would purchase item $j$ given that only the subset of substitutes for $i$ is available for purchase. In other words the customers probability of buying item $j$ given item $i$ is unavailable will remain independent from the different combination of substitutes of item $i$ present on the shelf. \newline
\emph {The reduction in specificity brought by the combination of assumption 1, 2, and 3 makes our proposed method scalable.}

\subsection{Data Preparation}
Expanding on our first assumption, we restrict our attention to modeling the item--item DT coefficients only, as there is no correlation between DT coefficient values and the values of the no-purchase probabilities for each item. Therefore, we leverage historical sales data, which does not contain information on no-purchase signals and includes only cases where customers decided to make a purchase.

We first take the transaction data and aggregate it to the store--item--week level. The result is a matrix $Z \in \mathbb{N}^{T \times n}$, where $T$ is the number of store-weeks in our transaction history, and $n$ is the number of items in the universe being considered. Each row $Z_i \in \mathbb{N}^n$ of this matrix represents the purchase counts of each item in our universe during the corresponding store-week $i$.

We then use out-of-stock (OOS) data and historical assortment data to create a matrix $W \in \{0,1\}^{T \times n}$ (note that this has the same shape as our purchase matrix $Z$), where each row $W_i \in \{0,1\}^n$ is a vector of binary indicators. Each element $W_{ij}$ is equal to 1 if and only if item $j$ was both: (1) on plan to be kept on the store shelf (part of the modular plan) for the category during store-week $i$, and (2) was actually available during store-week $i$. Constructing this requires joining the OOS and modular plan tables and taking the set intersection of the two.

Note that one can also construct the matrices $W$ and $Z$ at a per-purchase level (with $T$ representing the number of customers, rather than store-weeks), where each row represents the availability presented to and the purchase decision made by a single customer. In this case, the matrix $Z$ would be binary. Due to the form of the log-likelihood function in Abdallah and Vulcano [3], fitting the model using data constructed at the store-week level is equivalent to fitting it at the per-purchase level. We use the former in our implementation.

\subsection{Multinomial Logit (MNL) Model}

Multinomial Logit (MNL) models are a class of choice models [17] that have been widely used in econometrics and in revenue-management tasks such as SAO [3,6,14,18]. We follow the standard description of the MNL model given in Şimşek and Topaloğlu [14].

MNL models are popular largely because they rely on the previously mentioned IIA assumption [6]. This assumption greatly simplifies the statistical modeling problem by stating that the ratio of the log-odds of purchasing any two items is independent of the availability of other items [17]. In practice, this allows us to treat the purchase probabilities of items, given an offer set, as a function of a fixed parameter vector of item \emph{utilities}. This function takes the form of a softmax transformation [14,17].

In the MNL model, the mean utility parameter of an item $i \in U$ is $\eta_i$. If we offer a subset $O \subseteq U$ of items, then the probability that a customer purchases item $i$ is
\[
    \frac{e^{\eta_i}}{\sum_{j \in O} e^{\eta_j}}.
\]

The likelihood function corresponding to the parameter vector $\eta = (\eta_i)_{i \in U}$ is given in Abdallah and Vulcano [3]:
\[
    \mathcal{L}(\eta) 
    = \sum_{j=1}^n K_j \eta_j 
      - \sum_{t=1}^T m_t 
      \log\left( \sum_{i \in S_t} \exp(\eta_i) \right),
\]
where $m_t = \sum_{i=1}^n Z_{it}$ represents the total number of purchases in period $t$ (or a single purchase if the data is generated at a per-purchase level, in which case $m_t = 1$), and $K_j = \sum_{t=1}^T Z_{jt}$ represents the total number of purchases of item $j \in U$ over the observed historical period. One can efficiently maximize this function using the minorization--maximization method described in Abdallah and Vulcano [3].

Logit Models (LM) such as the MNL, despite their widespread use have two important drawbacks which hinders their usefulness for our specific optimization approach, especially at the scale at which we operate \newline

Firstly with respect to computing DT coefficients, the output of a random utility model such as the MNL consists in a vector of item preference utilities. These utilities can be used to determine, given an offer set $O \subseteq U$, the probability distribution of customer purchase over all of the items in $O$. However, the output of a choice model yields only a \textit{marginal} distribution of probabilities with respect to an item’s need state, since the conditional output of a choice model is conditional only on the availability of items on offer [6]. In particular, a customer's initial item preference determines the particular \textit{need state} they are trying to fulfill, which will affect the set of items they are willing to consider for purchase in the items place, if the initial item happens to be off-shelf. \newline

Therefore, the need state of an item needs to factor into the choice modeling analysis. Prior work has typically addressed this issue by performing analysis itself at an item hierarchy level that includes only items fulfilling the same need state. Vulcano et al [18], for example, consider a set of flights all serving the same Origin--Destination pair. A natural implementation of this strategy in our setting would be to fit a separate MNL model for each \textit{disjoint} group of items fulfilling a single need state. Then, for each item $i$, set the DT estimate
\[
\hat{\rho}_{ij} = \frac{\exp \hat{\eta}^G_j}{\sum_{k \in S \setminus \{i\}} \exp \hat{\eta}^G_k}, \quad \forall j \in G,
\]
and
\[
\hat{\rho}_{ij} = 0, \quad \forall j \notin G,
\]
where $\hat{\eta}^G$ is the estimated parameter vector of the model trained on the partition $G \subseteq U$ that contains $i$.

For our use case, this solution was impractical, because existing business hierarchy levels (1) are too numerous to render computation practical, (2) fail to either directly separate items by need-state, or (3) fail to satisfy the needed mutual disjointness property to keep the analysis well-defined. Thus, in our solution, we perform the analysis at a category level, which allows us to fit more precise models to item sub-universes instead of a general model fit to all categories (which would be inaccurate), but also does not yield an impractical number of choice models to be estimated. However, since item categories contain multiple need states which can overlap (are \textit{not} disjoint), we leverage our item substitution scores.  \newline

We then use the category-level utility estimates and leverage the IIA condition to allow us to restrict attention only to highly-substitutable (with substitution score \(> 0.6\)) items during inference, which is enabled by assumptions 1,2, and 3 above. The IIA condition allows us to use MNL parameters to construct a valid probability distribution over the substitutes of the deleted item $i$ using our category-level utilities, and interpret that as the probability that an arbitrary customer purchases any of the substitutes given that the substitutes are the only ones on offer. \newline

Finally, because we assume all DT coefficients across different need states are zero assumption 3, a Markov Chain model's transition probabilities for the items in a particular need state could be entirely characterized by its submatrix for that need-state. Furthermore, because of assumption 2, the rows of this matrix would be identical because the transition probabilities from any item state $i$ is fully determined by $i$'s need-state. So the matrix would have rank one, and thus as established in Theorem 3.1 of Blanchet et al.\ [5], we can interpret these MNL probabilities as transition probabilities, i.e.\ Demand Transfer coefficients. \newline

Our newly proposed modified approach enables us to avoid the pitfalls of both having to artificially partition a category into subsets not precisely representing correct need states, as well incur the computational cost by training many independent models for each partition.
We therefore call our improved algorithm a \textit{Restricted} Logit model, emphasizing the additional constraints to our inference we impose via the use of substitution. We discuss more details of our modified approach in the subsection Restricted Logit Model.

\subsection{Restricted Logit Model}

Having computed our substitute scores $\sigma \in [-1,1]$ for every item pair in our universe, we now have an indicator, for each item $i$, whether or not another item $j$ should be considered a candidate for demand transfer. Recall that fundamentally, a positive substitution indicator between a pair of items $i, j$ indicates that if one of the items in the pair is deleted, then the other should be considered a candidate for demand transfer. Otherwise, $\rho_{ij} = 0$.

The discussion in subsection "Multinomial Logit (MNL) Model" presents a natural way of incorporating our need-state information, as determined by our substitution indicators, into the Demand Transfer coefficient calculation of our Restricted Logit Model. This means, that given utility estimates $\hat{\eta} = (\hat{\eta}_n)_{n \in U}$, the final DT coefficient estimates for any given $i \in U$ should be given by:

\begin{equation}
\hat{\rho}_{ij} = \frac{\exp \hat{\eta}_j}{\sum_{k \in S \setminus \{i\}} \exp \hat{\eta}_k},
\quad \forall j \in S
\end{equation}

\begin{equation}
\hat{\rho}_{ij} = 0, \quad \forall j \in U \setminus S
\end{equation}

\begin{equation}
\hat{\rho}_{ii} = 0, \quad \forall i \in U
\end{equation}

where $S \subseteq U$ is the set of all substitutes for $i$.

We implemented the MNL fitting procedure described in (Abdallah and Vulcano [3]) for our Restricted Logit model across all categories using NumPy [8] and parallelized this algorithm across categories using Spark [15], parallelized across categories using internal distribute system.

\section{RESULTS AND FINDINGS}

To evaluate our model we perform offline backtesting on historical transaction data for items across a representative sample of locations for multiple categories. 

Logic:
\begin{enumerate}
    \item Post training the model on one year of transaction data, we keep the last two months of data as a test set.
    \item For this test set, we source the forecast and the actual sales observed for each item.
    
    \item We compute two WMAPE metrics:
    \begin{enumerate}
        \item \textbf{Forecast MAPE}:
       \[
	\text{MAPE}_{\text{forecast}}
	= \sum \frac{\left| \text{Forecast} - \text{Actual} \right|}{\left| \text{Actual} \right|}
	\]
        
        \item \textbf{Adjusted MAPE}:
  
	\[
	\text{MAPE}_{\text{adjusted}}
	= \sum \frac{\left| \text{Adjusted Demand} - \text{Actual} \right|}{\left| \text{Actual} \right|}
	\]
        \item \textbf{Adjusted WMAPE weighted against the actual units sold for the item}:
	\[
	\text{WMAPE}
	= \frac{\sum (\text{Actual} \cdot \text{MAPE})}{\sum (\text{Actual})}
	\]
	
        \item The adjusted demand for item $i$ is defined as
        \[
        \tilde{D}_i
        = \hat{D}_i
        + \rho_{ji}\,\hat{D}_j
        + \rho_{ki}\,\hat{D}_k,
        \]
        where $\hat{D}_i$, $\hat{D}_j$, and $\hat{D}_k$ denote the forecasted demands of items
        $i$, $j$, and $k$, respectively.
        
        Given items $\{i, j, k\}$ form a subset $S \subseteq U$ of the item universe and are assumed
        to be mutually substitutable.
        
    \end{enumerate}
\end{enumerate}

\begin{table}[!t]
\centering
\caption{Forecast vs Adjusted WMAPE Across Locations}
\label{table : WMAPE Comparison}
\footnotesize
\begin{tabular}{|c|c|c|c|}
\hline
\textbf{Location} & \textbf{Forecast WMAPE} & \textbf{Adjusted WMAPE} & \textbf{Difference} \\
\hline
A & 0.28 & \textbf{0.21} & -0.07 \\
B & 0.16 & \textbf{0.10} & -0.06 \\
C & 0.02 & \textbf{0.03} & +0.01 \\
D & 0.16 & \textbf{0.08} &  -0.08 \\
E & 0.18 & \textbf{0.07} & -0.11 \\
\hline
\end{tabular}
\end{table}

Table~\ref{table : WMAPE Comparison} reports offline backtesting results for a representative sample of locations across 50 categories (selected to span consumables and general merchandise with varying demand characteristics). The general trend observed was that appreciable reductions in forecast error occurred more frequently at locations with a higher baseline wMAPE, supporting our hypothesis that a significant portion of the raw forecast error is attributable to unaccounted demand transfer.

Across our test categories we observed consistent reductions in wMAPE, supporting the efficacy of DT correction, which helps to establish the efficacy of our approach to DT calculation at scale. This is particularly important for our use case, where we need to estimate DT coefficients at scale for a large number of items in a large number of categories.

Because actual customer substitution decisions are not directly observable in transaction data, forecast improvement serves as a proxy validation metric for demand-transfer estimation. Future work will explore customer-level substitution tracking to enable more direct validation of estimated coefficients.

\section{CONCLUSION}
In this paper, we introduced and described our Restricted Logit Model pipeline which is a modification of the original MNL model proposed by Abdallah and Vulcano [3]. Our Restricted Logit Model pipeline is created to solve the problem of estimating Demand Transfer at scale when used in conjunction with the proposed substitution framework. Our method is particularly useful in cases:
\begin{enumerate}
    \item Where it is known that the underlying need state of the items in the universe can differ. 
    \item Where item need-states are difficult to characterize.
    \item Need states do not correspond perfectly with a set of disjoint partitions of the item universe. 
    \item Where the item universe is huge having more than millions of items.
\end{enumerate}

To the best of our knowledge, few existing choice-model-based approaches simultaneously provide explicit demand-transfer coefficients, support overlapping need states, and remain computationally tractable for item universes of the scale considered in this work.
While the proposed framework has promising results, several opportunities remain for future research. In particular, relaxing the Independence of Irrelevant Alternatives (IIA) assumption through more flexible choice models, such as Nested Logit or Mixed Logit formulations, may better capture substitution behavior while preserving scalability. Further benchmarking against alternative demand transfer methods, including unrestricted MNL models, Markov Chain choice models, and modern machine learning approaches, as well as component-level ablation studies, would help provide a more comprehensive evaluation of the framework's effectiveness.

\section{Acknowledgement}
The authors would like to acknowledge Avinash Thangali for helpful discussions, contribution, and support.

\end{document}